\pdfoutput=1
\documentclass[11pt]{article}
\usepackage{acl}
\usepackage{microtype}
\usepackage{times}
\usepackage{latexsym}
\usepackage{xparse}
\usepackage{etoolbox}
\usepackage[utf8]{inputenc}
\usepackage{microtype}
\usepackage[T1]{fontenc}
\usepackage{multicol}
\usepackage{hyperref}
\usepackage{url}       
\usepackage{multirow}
\usepackage{float}
\usepackage{amsfonts}   
\usepackage{amsmath}
\usepackage{bbm}
\usepackage{nicefrac}   
\usepackage{microtype}  
\usepackage{graphicx}
\usepackage{xcolor}
\usepackage{caption}
\usepackage{booktabs,arydshln}
\usepackage{float}
\usepackage{blindtext}
\usepackage{multicol}
\usepackage{dblfloatfix}
\usepackage{linguex}
\usepackage{appendix}
\usepackage{mathtools}
\usepackage{enumitem}
\usepackage{xspace}
\usepackage{listings}
\usepackage{array,booktabs,longtable,tabularx,tabulary}
\usepackage{ltablex}
\usepackage[compact]{titlesec}

\makeatletter
\def\adl@drawiv#1#2#3{%
        \hskip.5\tabcolsep
        \xleaders#3{#2.5\@tempdimb #1{1}#2.5\@tempdimb}%
                #2\z@ plus1fil minus1fil\relax
        \hskip.5\tabcolsep}
\newcommand{\cdashlinelr}[1]{%
  \noalign{\vskip\aboverulesep
           \global\let\@dashdrawstore\adl@draw
           \global\let\adl@draw\adl@drawiv}
  \cdashline{#1}
  \noalign{\global\let\adl@draw\@dashdrawstore
           \vskip\belowrulesep}}
\makeatother

\restylefloat{table}
\usepackage{etoolbox}
\newcommand{\zerodisplayskips}{%
  \setlength{\abovedisplayskip}{4pt}%
  \setlength{\belowdisplayskip}{4pt}%
  \setlength{\abovedisplayshortskip}{4pt}%
  \setlength{\belowdisplayshortskip}{4pt}}
\appto{\normalsize}{\zerodisplayskips}
\appto{\small}{\zerodisplayskips}
\appto{\footnotesize}{\zerodisplayskips}

\newcommand{\cell}[1]{\begin{tabular}{@{}c@{}}#1\end{tabular}}

\newcommand{\smallsec}[1]{\vspace{-0.2em}\paragraph{#1.}}

\newcommand{\rephrasingtest}{\textsc{rephrase-inv}\xspace}
\newcommand{\rephrasingtestfull}{rephrasing invariance\xspace}
\newcommand{\rephrasingtestfullcap}{Rephrasing invariance\xspace}
\newcommand{\ordertest}{\textsc{order-inv}\xspace}
\newcommand{\ordertestfull}{order invariance\xspace}
\newcommand{\ordertestfullcap}{Order invariance\xspace}
\newcommand{\ontologicaltest}{\textsc{ontological-inv}\xspace}

\newcommand{\ontologicaltestfullcap}{Ontological invariance\xspace}
\newcommand{\visualobfuscationtest}{\textsc{visual-inv}\xspace}

\newcommand{\visualobfuscationtestfullcap}{Visual obfuscation invariance\xspace}
\newcommand{\negationtest}{\textsc{negation-dir}\xspace}

\newcommand{\negationtestfullcap}{Negation directional expectation\xspace}
\newcommand{\attributeantonymtest}{\textsc{antonym-dir}\xspace}

\newcommand{\attributeantonymtestfullcap}{Attribute antonym directional expectation\xspace}

\newcommand{\accuracy}{\textsc{acc}\xspace}
\newcommand{\accuracyfull}{accuracy\xspace}
\newcommand{\accuracyfullcap}{Accuracy\xspace}
\newcommand{\consistency}{\textsc{cons}\xspace}
\newcommand{\consistencyfull}{self-consistency\xspace}
\newcommand{\consistencyfullcap}{Self-consistency\xspace}
\newcommand{\compaccuracy}{\textsc{c-acc}\xspace}
\newcommand{\compaccuracyfull}{comprehensive accuracy\xspace}
\newcommand{\compaccuracyfullcap}{Comprehensive accuracy\xspace}

\definecolor{green}{rgb}{0.0, 0.65, 0.0}

\title{CARETS: A Consistency And Robustness Evaluative Test Suite for VQA}

\author{\hspace{1mm}Carlos E. Jimenez\qquad Olga Russakovsky\qquad  Karthik Narasimhan\\ Princeton University\\
	\texttt{\{\href{mailto:carlosej@princeton.edu}{carlosej},olgarus,karthikn\}@princeton.edu} 
}

\date{}

\begin{document}
\maketitle

\begin{abstract}
We introduce CARETS, a systematic test suite to measure consistency and robustness of modern VQA models through a series of six fine-grained capability tests. In contrast to existing VQA test sets, CARETS features balanced question generation to create \textit{pairs} of instances to test models, with each pair focusing on a specific capability such as rephrasing, logical symmetry or image obfuscation. We evaluate six modern VQA systems on CARETS and identify several actionable weaknesses in model comprehension, especially with concepts such as negation, disjunction, or hypernym invariance. Interestingly, even the most sophisticated models are sensitive to aspects such as swapping the order of terms in a conjunction or changing the number of answer choices mentioned in the question. We release CARETS to be used as an extensible tool for evaluating multi-modal model robustness.\footnote{Source code, data, and additional resources may be found at \href{https://github.com/princeton-nlp/CARETS}{https://github.com/princeton-nlp/CARETS}}

\end{abstract}

\section{Introduction}
\label{sec:introduction}

\begin{figure*}[ht]
    \centering
    \includegraphics[width=\textwidth]{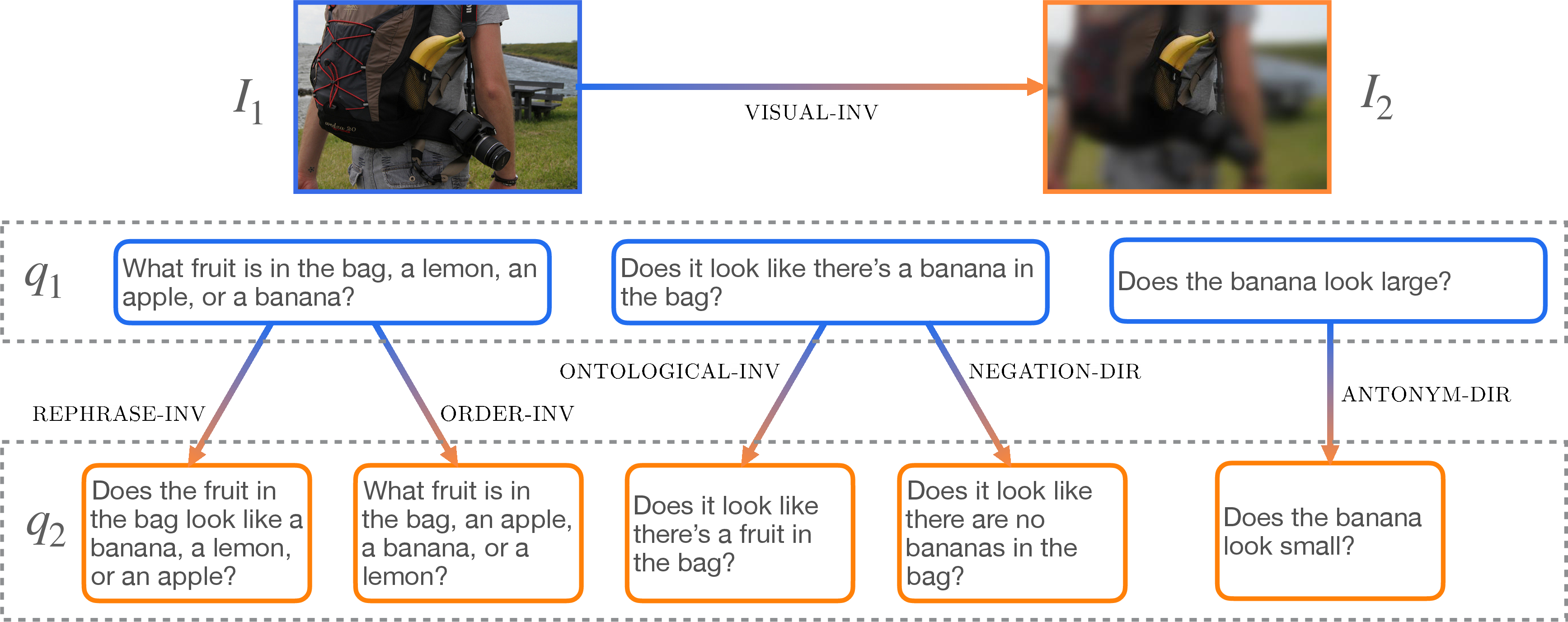}
    \caption{Our consistency and robustness test suite (CARETS) consists of six tests, corresponding to six identified phenomena that VQA models should be robust to. Tests evaluate models' predictions between pairs of instances (changing $q_1\rightarrow q_2$ or changing $I_1\rightarrow I_2$). \rephrasingtest for robustness to simple rephrasings; \ordertest for robustness to changed argument order in lists, conjunctions, and disjunctions; \ontologicaltest for understanding of ontology; \visualobfuscationtest for robustness to visual context perturbations; \negationtest for robustness to negative clauses; \attributeantonymtest for understanding of mutually exclusive attributes.}
    \label{fig:feature_figure}
\end{figure*}

The task of visual question answering integrates the domains of computer vision and NLP by probing models' understanding of images through natural language queries. After the introduction of the Visual Question Answering (VQA) benchmark \cite{VQA}, subsequent work identified the presence of several superficial correlations and other weaknesses latent in the VQA question gathering process \cite{vqa2, Agrawal2018}, which lead to potentially optimistic evaluations when considering accuracy alone. More recently developed benchmarks \cite{Hudson2019, vqa2, Agrawal2018}  explicitly avoid these weaknesses by introducing question, answer, and image balancing, or distributional shifts. While these efforts provide more difficult benchmarks, a thorough evaluation of model capabilities requires a deeper and more detailed approach.

To this end, we introduce CARETS -- a Consistency And Robustness Evaluative Test Suite for visual question answering. Inspired by recent work in NLP that generates `unit tests' for models~\cite{checklist}, CARETS contains systematically generated VQA tests that evaluate six different capabilities that any VQA model should handle -- robustness to question rephrasings, ontological reasoning, symmetrical logic, visual perturbations, question negation, and attribute antonymy. Each test point in CARETS consists of a \textit{pair} of instances which are small but strategic variations of each other either visually or in the question's text. This allows us to conduct fine-grained capability evaluations beyond just measuring high-level \accuracyfull scores. 

Across tests, we generate over 190,000 instance pairs in total using a programmatic approach that fills in templates (from nearly 200 templates in total) using ground-truth scene graphs~\cite{krishnavisualgenome} from the GQA~\cite{Hudson2019} validation split. We then evaluate six modern VQA models on each test using metrics of overall \accuracyfull, \textit{\consistencyfull} and \textit{\compaccuracyfull}. \consistencyfullcap measures models' ability to maintain their answer across question variations, while \compaccuracyfull estimates their ability to answer all instance variants correctly.

Our experiments reveal several interesting findings: (1) most modern VQA systems achieve only middling self-consistency ($\sim$60-80\%) which is further not always correlated with their \accuracyfull,  (2) all models struggle to comprehend the concept of negation (self-consistency of 18-28\% and \compaccuracyfull $<$17\%), and (3) even simple perturbations like changing the order of choices in the question text can induce a substantial drop (10-15\%) in performance. Moreover, even state-of-the-art models like LXMERT~\cite{Tan2020} are highly sensitive to the type of questions (binary vs multi-choice) and the number of choices provided. These results reveal several shortcomings in modern VQA systems and hint at potential areas for future improvement. Going beyond our current discoveries, CARETS is an extensible framework and may be easily extended by adding new capability tests for fine-grained evaluation of future models.

\section{Related Work}
\label{sec:related}

\smallsec{VQA evaluation}
The textual, visual, and answer biases discovered in the VQA dataset \cite{VQA} spurred on recent work seeking to improve model evaluation for the task by eliminating these biases \cite{vqa2}, introducing distributional shifts \cite{Agrawal2018}, requiring model explanations~\cite{VQAE}, thoroughly analyzing biases in datasets and models~\cite{ManjunathaCVPR2019}, or evaluating on different recognition subtasks~\cite{TDIUC}. While debiased and challenging benchmarks are important, their focus on accuracy as the sole evaluation metric leaves much to be desired \cite{Ribeiro2020, kervadec2020roses}. In contrast, our testbed provides question or image \textit{pairs} that compares models' predictions \emph{between} questions; measuring their accuracy, consistency, and robustness to a variety of text and image perturbations. 

\smallsec{Synthetic Dataset Generation for VQA} One way in which we can generate diverse and balanced datasets is to generate them synthetically, as is done by \cite{Johnson2015,Zhang2016YinAY,Hudson2019}. Synthetically generating questions, images, or both, allows fine control over the distribution of questions, answers, and images. Additionally for our case, synthetic generation allows us to control not just the particular semantics of one question, but also how one question relates to another question in a precisely defined way (e.g. one question is a negation of another) while also remaining relevant and grounded in the image. As both the CLEVR \cite{Johnson2015} and GQA \cite{Hudson2019} datasets use image scene graphs for question and label generation, they contain questions combining a variety of required capabilities, including compositionality. We feature real-world images with synthetically generated questions as well, but in contrast to GQA, our evaluation has instance \textit{pairs} to systematically test a focused set of capabilities, showing that models may still struggle with simpler, non-compositional questions. 

\smallsec{Consistency as Model Comprehension} Some recent work has sought to evaluate models using consistency and other metrics \cite{Hudson2019, Shah2019, Ribeiro2020, Selvaraju2020SQuINTingAV, Bitton2021AutomaticGO}. These evaluations often evaluate consistency through question entailment and implication, or simply contrasting examples in the case of \cite{Bitton2021AutomaticGO}. While we consider such methods important for evaluating model comprehension, they often combine question types and capabilities, changing the kind of expected answer, or evaluating consistency on a tree or set of entailed questions. While models would ideally be consistent and robust for these more complex types of tests, our approach reveals that models can fail even on simple implications.


\section{Fine-grained capability tests}



Our goal is to provide a testbed for fine-grained evaluation of VQA models' capabilities. To do so, we generate multiple tests, each corresponding to a specific model capability. In contrast to standard VQA benchmarks~\cite{VQA,vqa2,Hudson2019}, our test sets consist of a \emph{pair} of \emph{original} and \emph{perturbed} instances $\langle (I_1, q_1, a_1), (I_2, q_2, a_2) \rangle$, each with an image, a question and an answer. The two instances within a pair differ from each other in a minimal yet carefully constructed way to hone in on a particular capability, similar to BLiMP \cite{blimp}.
A model is then evaluated on its overall \emph{\accuracyfull}, \emph{self-consistency} (ability to produce consistent, even if wrong, answers to both instances within a pair), and \emph{\compaccuracyfull} (ability to answer consistently and correctly for an instance pair).

Overall, we create six tests corresponding to key capabilities. We provide a high-level description of each test here and describe generation details in Section~\ref{sec:dataset}. First, we create four \emph{invariance} tests\footnote{Reusing terminology from \citet{checklist}.} that use variations of the question phrasing and expect the model to produce the \emph{same} answer to both questions within an instance pair: 
\begin{enumerate}[noitemsep,topsep=2pt]
    \item {\bf \rephrasingtestfullcap} (\rephrasingtest) measures the model's understanding of minor, meaning-preserving textual changes, e.g.: ``What color is the bottle on the shelf, white or blue?'' and ``\textit{Does the} color of the bottle on the shelf \textit{seem more} white or blue?''
    \item {\bf \ontologicaltestfullcap} (\ontologicaltest) measures understanding of ontology, e.g. changing a hyponym in: ``Do you see a green jacket?'' to a hypernym ``Do you see any green \emph{clothes}?''
    \item {\bf \ordertestfullcap} (\ordertest) measures understanding of logically equivalent questions containing different argument orders, e.g.: ``Is the black vehicle a van or a truck?'' and ``Is the black vehicle a \emph{truck} or a \emph{van}?''
    \item {\bf \visualobfuscationtestfullcap} (\visualobfuscationtest) measures the model's answering ability when parts of the image not directly relevant to the visual question are removed. Specifically, we explore blurring, masking and cropping techniques to modify the image.
\end{enumerate}

We also create \emph{directional expectation} tests to measure model behavior on instance pairs where the answer is expected to \emph{change}:
\begin{enumerate}[resume,noitemsep,topsep=2pt]
    \item {\bf \attributeantonymtestfullcap} (\attributeantonymtest) measures the model's understanding of antonyms, e.g., ``Do you think that the wood table is short?'' and ``Do you think that the wood table is \emph{long}?''
    \item {\bf \negationtestfullcap} (\negationtest) measures a model's grasp of negation,  e.g., `` Are there any apples in this picture?'' and ``Are there \emph{no} apples in this picture?
\end{enumerate}


\section{Dataset generation}

\label{sec:dataset}

Each of the six test datasets start with the generation of `original', unperturbed instances $(I_1, q_1, a_1)$ (Section~\ref{sec:dataset_origq}). Then, for each such instance, we generate a variation $(I_2, q_2, a_2)$ by either perturbing the original question $q_1$ or image $I_1$ (Section~\ref{sec:dataset_newq}). Further, each test set is composed of a diverse set of questions. These may broadly be grouped into \textit{verification} (or \textit{binary}) questions, with expected answers being either \textit{yes} or \textit{no}, and \textit{multi-choice} questions, with expected answers derived from a list of objects or attributes provided in the question.

\subsection{Original instance generation}
\label{sec:dataset_origq}

Questions for each test are generated from question templates (examples for each are provided in Appendix \ref{app:test_dataset_statistics_section}) grouped into the following types. 
\begin{itemize}[noitemsep,topsep=2pt]
    \item [Q1:] {\bf Object verification} (54 templates): e.g., ``Is there a \texttt{<obj>} in the image?'' 
    \item [Q2:] {\bf  Conjunctive verification} (18 templates): e.g., ``Is there a \texttt{<obj1>} and a \texttt{<obj2>} in the image?''
    \item [Q3:] {\bf Disjunctive verification} (18 templates): e.g., ``Is there a \texttt{<obj1>} or a \texttt{<obj2>} in the image?''
    \item [Q4:] {\bf Attribute verification} (25 templates): e.g., ``Is the \texttt{<obj>} in the image \texttt{<attr>}?''
    \item [Q5:] {\bf Object multi-choice} (25 templates): e.g., ``Is the \texttt{<obj-class>}, \texttt{<choices>}?''
    \item [Q6:] {\bf Attribute multi-choice} (39 templates): e.g., ``What sort of \texttt{<attr-class>} is the \texttt{<obj>}, \texttt{<choices>}?''
    \item [Q7:] {\bf Action multi-choice} (28 templates): e.g., ``What is the \texttt{<action-}\texttt{class>} that the \texttt{<obj>} doing, \texttt{<choices>}?''
\end{itemize}
Words in \texttt{typewriter} font represent template arguments. Generally, each \texttt{<obj>} argument can be filled by a singular object (``cup'') or an attribute+object (``red cup'') while \texttt{<attr>} arguments are filled with singular attributes (``shiny''). For object verification (Q1), attribute verification (Q4),  attribute multi-choice (Q6), and action multi-choice (Q7) questions, some templates let \texttt{<obj>} arguments be filled with an object related to another object (e.g. ``red cup \textit{on the} table''); this type is excluded from conjunctive verification (Q2) and disjunctive verification (Q3) questions to prevent the generation of verbose questions. 

For the \textbf{multi-choice} questions (Q5, Q6, Q7), \texttt{<choices>} are replaced with a list of 2 or 3 \textit{singular} objects, attributes, or actions respectively (e.g. ``cow or horse'' or ``wood, metal, or plastic''). The \texttt{<obj-class>} argument is filled with a hypernym of all object choices and always appears with either an attribute or a related object (``black animal'', ``animal \textit{eating} grass''). The \texttt{<attr-class>} argument is filled with the attribute category of all attribute choices (e.g. ``material''). Finally, the \texttt{<action-class>} argument is filled with the action category of all action choices (e.g. ``sport'').

\smallsec{Question argument generation} 
The question arguments are generated using images from the validation split of the GQA dataset \cite{Hudson2019} which contains 10,696 images manually annotated with 1,536 different objects and 603 different object attributes. 

To generate questions, we sample objects and attributes directly from an image's scene graph annotation to populate a question type's arguments. For \textit{binary} question types this results in questions with solely affirmative answers. To produce an answer balanced dataset, we run a second stage of question argument generation for \textit{binary} questions to generate \emph{plausible} negative questions with \textit{false} objects or attributes. We sample \textit{false} objects from a distribution conditioned on an image's objects,
and optionally sample object attributes from a distribution conditioned on the chosen object.

For \texttt{<choices>} arguments, \textit{false} choices are again generated from a distribution conditioned on the object's hypernym for Q5 questions, the attribute category for Q6 questions, or the action category for the Q7 questions. We additionally ensure that the generated choices are mutually exclusive (e.g. ``tan or beige'' would be an invalid generation). To get more diverse multi-choice questions, we first generate a large pool of question candidates, and then select only a small number of questions sampled from this pool with sample probabilities inversely proportional to the count of the questions' hypernym, attribute class, or action class, and the count of the generated answer.

\smallsec{Question argument refinement} To improve the reliability of generated questions, we apply a variety of checks and constraints. For example, when sampling \textit{false} objects from the conditional distribution, we filter out all objects (and their hypernyms\footnote{We use an ontology with hypernym paths generated with WordNet~\cite{wordnet}. We manually review and revise the default synset annotations from Visual Genome \cite{krishnavisualgenome} for the entire object vocabulary, and compare to a sample of annotated images to ensure general validity.}) present in the scene graph in order to guarantee that the sampled object is truly not present. We also filter out question arguments that are not included in the image scene graph but are sub-parts of objects that are annotated (e.g., ``tire'' when a ``car'' is annotated). Finally, we enforce various logical constraints on question arguments to prevent trivial or malformed questions. For example, for conjunctive and disjunctive questions (Q2, Q3), we apply a \emph{hypernym exclusion} constraint to prevent questions like ``Is there a \texttt{black cat} and a \texttt{black animal} in the image?''.

\subsection{Perturbed pair generation}
\label{sec:dataset_newq}


We now describe our procedure for creating perturbed instances $(I_2, q_2, a_2)$ for the six tests. In all tests except visual obfuscation, the image remains unchanged, i.e. $I_2 = I_1$.


\smallsec{(a) \rephrasingtestfullcap}  Since each original question $q_1$ was generated using a text template, we simply use a different template of the same type to generate a valid rephrasing $q_2$. The image and answer remain the same, i.e. $I_1 = I_2,a_1=a_2$ and the model is expected to be invariant to this rephrasing. We apply this to Q1, Q2, Q3, Q5, Q6 and Q7.




\smallsec{(b) Ontological invariance} Here, we use object verification questions (Q1) only and perform two types of transformations. For positive questions (i.e. $a_1$ = \emph{yes}), we filter question arguments to only include objects that are valid hyponyms (using WordNet again)
and use those to generate a perturbed question $q_2$ by changing the hyponym to a hypernym. For example, $q_1$ = ``Do you see a jogging \emph{woman}?'' with $a_1$ = \emph{yes} is paired with $q_2$ = ``Do you see a jogging \emph{person}?'' containing a hypernym. 
Similarly, for negative questions ($a_1$ = \emph{no}), we filter question arguments to only include valid hypernyms and generate a $q_2$: thus for example, $q_1$ = ``Do you see a jogging \emph{person}'' with $a_1$ = \emph{no} is paired with $q_2$ = ``Do you see a jogging \emph{woman}?'' containing a hyponym, $a_2$ = \emph{no} also. 

\smallsec{(c) \ordertestfullcap} \ordertestfullcap tests apply to conjunctive verification, disjunctive verification, and all multi-choice question types; models are expected to be invariant to the logical order of arguments.
We perturb conjunctive verification and disjunctive verification questions by swapping the questions' first and second arguments (\texttt{<obj1>, <obj2>}). For multi-choice question types, we perturb instances by generating the \texttt{<choices>} argument with different orders. The answer remains the same in both cases by construction.

\smallsec{(d) \visualobfuscationtestfullcap} For this test, we let $q_1=q_2$ and $a_1=a_2$ but generate a perturbed image $I_2$ by obscuring parts of $I_1$ that are irrelevant to the question at hand using bounding box annotations from Visual Genome~\cite{krishnavisualgenome}.  For all \textit{true} objects in a question, we consider the bounding boxes around these object(s) to be the \emph{foreground} and all other pixels in the image to be the \emph{background}. For negative verification questions asking about object(s) not present in the image, we select one (or two) random object bounding box(es) as the foreground and consider everything else to be the background, since focusing on any image region should not affect the model's answer.\footnote{We choose $32\times 32$ as a minimum bounding box size, shown to be reliably recognized by humans \cite{tiny_image}. } We then apply five types of perturbations to obscure the background: (i-iii) {\bf Gaussian blurring} using the soft masking method of~\cite{yang2021study} with light ($\sigma=3$), medium ($\sigma=6$), or heavy ($\sigma=9$) blurring, (iv) {\bf Masking} with the channel-wise averaged pixel value from the GQA \cite{Hudson2019} training dataset, entirely obscuring the context, and (v) {\bf Cropping} to the smallest rectangular region including the foreground. Example images are shown in Appendix \ref{app:visual_obfuscation_invariance}.

\smallsec{(e) \negationtestfullcap} For the negation directional test, we use object verification, conjunctive verification, and disjunctive verification questions. Each question $q_1$ is perturbed by substituting the original's text template with a paired negated text template to create $q_2$. Since each perturbed question represents the negation of the original, the expected answers $a_1 \neq a_2$.

\smallsec{(f) \attributeantonymtestfullcap} We perturb the generated attribute verification questions by changing the \texttt{<attr>} question argument to its antonym using WordNet.  All attribute antonym relations are manually curated to remove unintuitive examples; questions with arguments without a valid antonym are discarded. The original and perturbed questions of a pair have opposite answers $a_1 \neq a_2$. 

\section{Experimental Setup}
\label{sec:training_details}

\subsection{Human baseline}
\label{sec:human}

To assess the quality, difficulty and validity of the generated tests, we sample 100 question \textit{pairs} (200 questions) from each question type for the 6 tests and procure 5 annotations per question from workers on Amazon Mechanical Turk. Workers are vetted for a 97\% approval rating and minimum of 500 completed HITs. Workers take $\sim 2$ minutes per task on average and are compensated $\$0.50$ per task and thus $\sim \$15$ per hour. Each HIT include 24 questions total, including 4 verification questions\footnote{Tasks are also interspersed with binary or multi-choice ``gold-standard'' questions with perfect annotator agreement from the VQA dataset~\cite{VQA} which are required to be answered correctly before a HIT can be submitted. Workers are given the opportunity to correct answers before submitting if a ``gold-standard'' question has been answered incorrectly.}, and typically include a variety of question types from each of our tests. 

\smallsec{Human agreement} In addition to ``yes'' and ``no'' for binary questions and the appropriate choices for multi-choice questions, all questions offer an \textit{ambiguous} option. Human answers are the majority vote among the 5 workers; questions failing to reach majority or with \textit{ambiguous} as the majority are always counted against accuracy. This process is inspired by the human evaluations of implied question pairs in \citet{Ribeiro2020}. We report both human and model performance in Section \ref{sec:results}.

\subsection{Evaluated models}
\label{sec:models}
We evaluate six recent models on our tests, and compare them to human accuracy. Models are trained on the GQA~\cite{Hudson2019} balanced training split (using hyperparameters suggested from the original papers). All models, except LXMERT\footnote{For LXMERT we use the authors' open source repository at \href{https://github.com/airsplay/lxmert}{https://github.com/airsplay/lxmert}}, are trained and finetuned using the MMF \cite{singh2020mmf} library and region of interest (RoI) features from Faster R-CNN \cite{Ren2015FasterRT} with a ResNeXt-152 \cite{resnext} backbone pre-trained on the Visual Genome \cite{krishnavisualgenome} dataset for object-based models.
More details are provided in Appendix \ref{app:more_training_details_section}.

\smallsec{Model initialization and pre-training} Of the six models evaluated, a defining characteristic of each model relates to their initializations, image-encoding choice, and the use of multi-modal pre-training. Our most basic model (CNN+LSTM) is randomly initialized and uses no pre-trained components; however, GloVe \cite{pennington2014glove} word embeddings are used for representing tokens. Another class of models use pre-trained image encoders to extract object features from images. Of our models, BAN \cite{ban} is randomly initialized prior to training but ingests pre-trained Faster R-CNN features which should provide the model with enhanced visual capabilities over the CNN+LSTM model.  MultiModal BiTransformer (MMBT) \cite{kiela2019supervised}, uses similar pre-trained image features as BAN but is further initialized with pre-trained BERT \cite{Devlin2019BERTPO} weights prior to training on GQA. The last class of models are multi-modal pre-trained models; those that use pre-trained image features and are pre-trained on multi-modal tasks, such as image-based masked language modeling. Models in this class include LXMERT \cite{Tan2020}, ViLBERT \cite{Lu2019ViLBERTPT}, and VisualBERT \cite{li2019visualbert}. Similar to MMBT, ViLBERT and VisualBERT are also initialized with pre-trained weights from BERT. 

\subsection{Metrics}
\label{sec:metrics}
\textbf{\accuracyfullcap (\accuracy).} On our test datasets with $K$ paired instances, we define \accuracyfull as:
\[
\frac{1}{2K}\sum_{i=1}^K \mathbbm{1}[\hat{a}_1^i = a_1^i] +\mathbbm{1}[\hat{a}_2^i = a_2^i] 
\]
where the model answers $\hat{a}_a^i,\hat{a}_2^i$ on the original and perturbed questions respectively are compared to the 
 ground truth answers $a_1^i,a_2^i$.\\
\textbf{Self-consistency (\consistency).} We measure self-consistency of the model predictions across the original and perturbed questions as 
\begin{align*}
    \frac{1}{K}\sum_{i=1}^K \left \{ \begin{array}{ll}
    \mathbbm{1}[\hat{a}_1^i = \hat{a}_2^i] & \text{on invariance tests} \\
   \mathbbm{1}[\hat{a}_1^i \neq \hat{a}_2^i] & \text{on directional exp. tests} 
    \end{array} \right .
\end{align*}
Note that these metrics only measure the internal consistency of the model and do not include the ground truth answers $a_1^i,a_2^i$.\\
\textbf{\compaccuracyfullcap (\compaccuracy).} We define \compaccuracyfull as:
\[
\frac{1}{K}\sum_{i=1}^K \mathbbm{1}[\hat{a}_1^i = a_1^i \land \hat{a}_2^i = a_2^i] 
\]
measuring whether model predictions are both accurate and self-consistent across perturbations.

\section{Results}
\label{sec:results}

\newcommand{\figwidth}{19.5em}

\begin{figure*}[t]
\centering
\includegraphics[width=\textwidth]{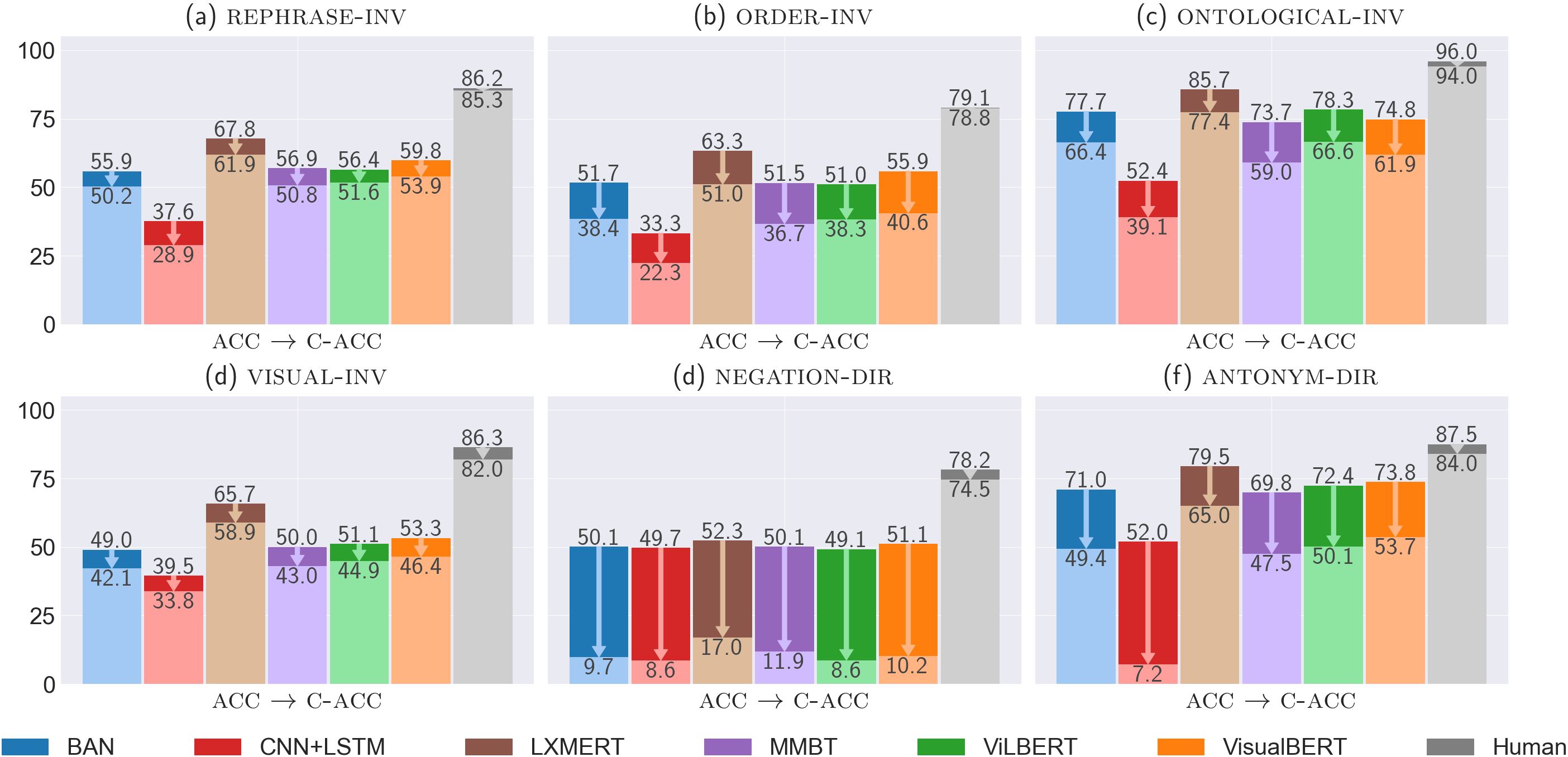}
\caption{\accuracy and \compaccuracy across all six tests. 
All models perform worse than humans and exhibit consistent drops in \compaccuracy from \accuracy, especially struggling on \negationtest and \attributeantonymtest tests. Best viewed in color.}

\label{fig:all_tests}
\vspace{-2mm}
\end{figure*}

\paragraph{(R1) Modern VQA models are not robust to invariance and directionality tests.}

Fig \ref{fig:all_tests} details the performance of various models under our suite of tests. Each bar in the figure shows both  \accuracy and \compaccuracy for the model, with the arrow representing the gap between the two. We first observe that all models achieve significantly lower performance (at least a 8\% drop) compared to humans (grey). Even simple tests such as \rephrasingtest (Figure~\ref{fig:all_tests}(a)) prove to be quite challenging, with models managing $<$ 68\% \accuracy compared to humans' 86\%. On tests like \negationtest (Figure~\ref{fig:all_tests}(e)), models only get about 50\% accuracy, substantially worse than human scores of 78\%.

Moreover, \compaccuracy is considerably lower than \accuracy across the board, with as much as a 35\% gap on \negationtest and 14.5\% on \attributeantonymtest tests, even for a state-of-the-art model like LXMERT.\footnote{Other modern systems like ViLBERT and VisualBERT have even worse \compaccuracy.} Even though this gap is smaller on other tests like \rephrasingtest or \visualobfuscationtest, the performance drop is still at least 6-7\% in most cases. This means that models are not invariant to textual rephrasings of questions and do not have a strong grasp of concepts like attributes and negation, despite negation of attributes appearing in the GQA training dataset.

\begin{table}[t]
\centering
\resizebox{0.49\textwidth}{!}{
\caption{\consistencyfullcap scores for all tests. Models only achieve middling \consistencyfull, compared to >89\% human \consistencyfull.}
\label{tab:full_consistency_table}
\begin{tabular}{l@{\hskip 0.00in}c@{\hskip 0.06in}c@{\hskip 0.06in}c@{\hskip 0.06in}c@{\hskip 0.06in}|c@{\hskip 0.06in}c@{\hskip 0.06in}}
\toprule
               &      \multicolumn{6}{c}{\consistency}  \\\cmidrule{2-7}
               &          Reph. &          Order &          Onto. &      Vis. &    Neg. &    Attr. \\
\midrule   
BAN            &           82.3 &           65.2 &           77.4 &                82.0 &           56.8 &           19.3 \\
CNN+LSTM       &           68.9 &           62.6 &           73.4 &                80.0 &           10.5 &           17.8 \\
LXMERT         &           83.8 &  \textbf{71.0} &  \textbf{83.5} &                78.6 &  \textbf{71.1} &  \textbf{28.1} \\
MMBT           &           81.7 &           58.6 &           70.5 &                81.4 &           54.6 &           23.1 \\
ViLBERT        &  \textbf{86.4} &           66.2 &           76.5 &                83.4 &           55.3 &           18.7 \\
VisualBERT     &           83.6 &           60.8 &           74.1 &       \textbf{83.8} &           59.8 &           18.3 \\\cdashlinelr{1-7}
Human          &           96.6 &           98.8 &           94.0 &                89.0 &           89.0 &           88.5 \\
\bottomrule
 \end{tabular}
 }
\end{table}

\paragraph{(R2) VQA systems are not self-consistent in their predictions.}
Table~\ref{tab:full_consistency_table} shows the \consistencyfull scores for all models under our different tests. While humans achieve \consistency > 88\% in all the tests, VQA models are much worse -- at least 6\% lower \consistency in all cases, with the best performing model (LXMERT) still 26\% lower than human performance on average across tests and models.
Scores are especially low on the directional tests (antonym and negation), which means that models are confused in their decisions simply with the addition of negation words -- this hints at issues of overfitting to spurious feature without understanding the presence or absence of specific concepts, corroborating the findings of \cite{Bitton2021AutomaticGO}. Interestingly, the best performing model (LXMERT) is not always the most consistent.
Furthermore, there is no single model that is the most self-consistent, with LXMERT, ViLBERT and VisualBERT each returning the highest consistency scores on different tests. 

\begin{table}[t]
\centering
\caption{Ontological invariance results breakdown: original vs perturbed \accuracyfull, and hyponym vs hypernym self-consistency.}
\label{tab:hypernym_hyponym_invariance}
\begin{tabular}{lcc|cc}
\toprule
            &  Orig. &  Pert. &  Hyper- &  Hypo- \\
\midrule
BAN            & \textbf{79} &          76 &          75 & \textbf{79} \\
CNN+LSTM       &          49 & \textbf{56} & \textbf{75} &          72 \\
LXMERT         & \textbf{89} &          82 &          80 & \textbf{87} \\
MMBT           & \textbf{82} &          66 &          61 & \textbf{80} \\
ViLBERT        & \textbf{85} &          72 &          69 & \textbf{84} \\
Visual BERT    & \textbf{81} &          69 &          68 & \textbf{80} \\\cdashlinelr{1-5}
Human          & \textbf{96} & \textbf{96} &          91 & \textbf{96} \\
\bottomrule
\end{tabular}
\end{table}

\paragraph{(R3) Models are more robust to hyponym than hypernym variations.}
Breaking out the results on the ontological invariance test (Figure~\ref{fig:all_tests} (c)) in the last two columns of Table~\ref{tab:hypernym_hyponym_invariance}, we see that self-consistency is higher on the hyponym perturbations (on negative answer questions) than on hypernym perturbations (positive questions); this effect is particularly noticeable for MMBT and ViLBERT with a 19\% and 15\% difference, respectively. Thus, when an object is not detected in an image its hyponym elicits a negative response as expected; however when an object (like ``steak'') is  detected, the hypernym question (``Is there any \emph{meat} in the image?'') may trip the model to generate a negative response. This points to the need for more structured, hierarchical grounding of concepts in these models.

\paragraph{(R4) Models perform better on conjunctive rather than disjunctive tests.}
From Table~\ref{tab:conj_dij_order_invariance}, we note that models generally have higher accuracy on conjunctive rather than disjunctive tests, with the largest discrepancy for LXMERT at 81\% accuracy on conjunctive tests vs only 62\% on disjunctive. Many models seem to exhibit a strong positive bias for disjunctive questions, suggesting they may just be short-cutting to answering `yes' for disjunctive questions. LXMERT also seems to frequently confuse disjunctive questions for an open-ended or multi-choice question.

\begin{table}[t]
\centering
\caption{Conjunctive (Con) vs disjunctive (Dis) \compaccuracyfull on \ordertest, along with a breakdown of response rates for yes (Y), no (N) and other than yes/no (O).}
\label{tab:conj_dij_order_invariance}
\begin{tabular}{l@{\hskip 0.02in}c@{\hskip 0.05in}c@{\hskip 0.05in}c@{\hskip 0.05in}c@{\hskip 0.1in}|c@{\hskip 0.05in}c@{\hskip 0.05in}c@{\hskip 0.05in}c}
\toprule
            & \multicolumn{4}{c|}{Conjunction} & \multicolumn{4}{c}{Disjunction} \\\cmidrule{2-9}
            &   \accuracy &           Y &           N &           O &   \accuracy &           Y &           N &           O \\\midrule
BAN         &          52 &          53 &          47 &           0 &          52 & \textbf{71} &          28 &           0 \\
CNN+LSTM    &          39 &          53 &          47 &           0 &          35 & \textbf{65} &          35 &           0 \\
LXMERT      &          78 &          49 &          51 &           0 &          56 & \textbf{59} &          32 &           9 \\
MMBT        &          56 &          50 &          50 &           0 &          55 & \textbf{63} &          34 &           2 \\
ViLBERT     &          58 &          54 &          46 &           0 &          56 & \textbf{79} &          21 &           0 \\
VisualBERT  &          59 &          49 &          51 &           0 &          57 & \textbf{70} &          30 &           0 \\
\bottomrule
\end{tabular}
\end{table}

\begin{table}[t]
\centering
\caption{Comparison of LXMERT's behavior for questions with binary answer types and multi-choice answers. Metrics are reported as averages over \rephrasingtestfull and \ordertestfull tests. Human \accuracy reports human performance across test sets.}
\label{tab:choice_tests}
\begin{tabular}{l@{\hskip 0.09in}c@{\hskip 0.09in}c@{\hskip 0.09in}c@{\hskip 0.09in}|c}
\toprule
            &       \accuracy &  \consistency &  \compaccuracy  & \cell{Human\\\accuracy}\\
\midrule
Binary         &    \textbf{74.5} &    \textbf{89.3} &   \textbf{69.8} &          82.8 \\
Multi-choice   &           59.0 &           69.7 &          47.3 &          83.6 \\
2-choice       &           62.4 &           70.8 &          49.9 &          83.4 \\
3-choice       &           55.4 &           68.4 &          44.5 &          83.9 \\
\bottomrule
\end{tabular}
\end{table}

\paragraph{(R5) Models are sensitive to answer types and the number of choices in a question.}
Table \ref{tab:choice_tests} provides a breakdown of LXMERT's scores for binary and multi-choice questions. It is evident that multi-choice questions are harder for the model, with self-consistency dropping by 16\% between binary and multi-choice questions, and \compaccuracy dropping by 33\%. This is surprising since the multi-choice questions only include two or three choices and hence are quite similar to the binary (yes/no) questions. This may indicate a bias in the models towards binary questions with simple answers. Furthermore, Table \ref{tab:choice_tests} also shows that models consistently perform worse on 3-choice questions than 2-choice ones, with even the top-performing LXMERT having a 7\% drop from 62\% to 55\%. This hints that there may be some effect of randomness in the way these models pick their answers. In contrast and as expected, humans are robust to the number of choices.

\begin{table}[ht!]
\centering
\caption{
We augment the LXMERT model by adding 95,000 questions (9.9\% of the training dataset) to the training dataset and keep all other training procedures the same. The added questions are generated using the same methodology as CARETS but using images from the GQA training set. GQA validation accuracy stays comparable at 70.60\% for LXMERT (Augmented) vs. 70.67\% for LXMERT (GQA).
}
\label{tab:augmented_results_comparison_table}
\resizebox{0.49\textwidth}{!}{
\begin{tabular}{l@{\hskip 0.07in}c@{\hskip 0.07in}c@{\hskip 0.07in}c@{\hskip 0.07in}}
\toprule
                &  \multicolumn{3}{c}{LXMERT (Augmented)} \\\cmidrule{2-4}
                &  \accuracy & \consistency &  \cell{\compaccuracy} \\
\midrule
\rephrasingtest       & 84.0 \color{green}{$\uparrow 16.2$} & 94.1 \color{green}{$\uparrow 10.3$} & 81.5 \color{green}{$\uparrow 19.6$}\\
\ordertest            & 82.1 \color{green}{$\uparrow 18.8$} & 92.3 \color{green}{$\uparrow 21.3$} & 78.8 \color{green}{$\uparrow 27.8$}\\
\ontologicaltest      & 95.3 \color{green}{$\uparrow$ \hspace{0.5em}9.6} & 94.0 \color{green}{$\uparrow 10.5$} & 92.4 \color{green}{$\uparrow 14.9$}\\
\negationtest         & 88.3 \color{green}{$\uparrow 36.0$} & 90.5 \color{green}{$\uparrow 62.4$} & 83.6 \color{green}{$\uparrow 66.6$}\\
\attributeantonymtest & 65.0 \color{green}{$\uparrow$ \hspace{0.5em}5.5} & 83.5 \color{green}{$\uparrow 12.4$} & 76.7 \color{green}{$\uparrow 11.6$}\\
\bottomrule
\end{tabular}
}
\end{table}

\paragraph{(R6) Visual perturbations are easier for models to deal with.}
From Figure~\ref{fig:all_tests} and Table~\ref{tab:full_consistency_table}, we notice that the models are slightly more robust to visual perturbations on average compared to the lexical ones. All models only have a drop of 4-8\% from \accuracy to \compaccuracy, while self-consistency of all models is also 78\% or higher. Appendix \ref{app:visual_obfuscation_invariance} provides a more detailed breakdown of all the different visual perturbation tests we performed. 

\paragraph{(R7) Direct data augmentation improves CARETS evaluation.}
We show the feasibility of high performance on CARETS through data augmentation. We add 95,000 questions generated from CARETS question templates, and using a similar distribution of question types, to the original training split of GQA and re-train the LXMERT model. Table \ref{tab:augmented_results_comparison_table} shows that this dramatically improves the model on all three metrics (\accuracy, self-consistency and \compaccuracy), with the LXMERT(Augmented) model achieving near human performance on tests like \ordertest and \attributeantonymtest. Since CARETS is designed to be an evaluation suite, these numbers show that CARETS questions should generally be within the capabilities of existing SOTA models, provided that they are able to generalize appropriately.

\section{Conclusion \& Future Work}
\label{sec:conclusion}

In this work, we have developed CARETS -- a new test suite for capability-focused robustness testing and comprehension evaluation of visual question answering (VQA) models. CARETS consists of six different tests that use instance \textit{pairs} to evaluate models on their understanding of various linguistic and visual concepts. Using this test suite, we evaluated six modern VQA systems to reveal several inconsistencies in state-of-the-art models and provide a fine-grained view of their comprehension of different visuo-linguistic concepts. Quite surprisingly, we find that even state-of-the-art models struggle with concepts like negation, disjunction, order invariance and multi-choice questions.  CARETS can also support the addition of more tests in the future and we view it as a platform for continually stress testing and improving VQA models. 

CARETS emulates previous work in using text templates to generate questions and their textual perturbations~\cite{Hudson2019,Johnson2017,checklist}. The use of templates to generate perturbations is motivated by the desire to maintain the grounded integrity of generations, ensuring that they remain relevant and that the generated label is true in the context of the subject image. While we have sought to generate a diverse language set through using a large number of templates (nearly 200 in total), there are some limitations to this approach. An improvement to our approach may be able to generate more sophisticated questions and perturbations through conditional text generation~\cite{Schick2021GeneratingDW, Madaan2021GenerateYC, polyjuice} while also preserving our other motivations, such as atomicity and grounded relevancy.

\section*{Acknowledgments}
This material is based upon work
supported by the National Science Foundation under Grant
No. 2107048. Any opinions, findings, and conclusions or
recommendations expressed in this material are those of the
author(s) and do not necessarily reflect the views of the National Science Foundation.

\bibliographystyle{acl_natbib}
\bibliography{main}

\appendix
\section{Appendix}
\label{appendix:a}

\subsection{Test Dataset Examples}
\label{app:test_dataset_statistics_section}
We provide examples of templates for each question type here. In Section \ref{sec:dataset_origq}, we simplified the template arguments for readability. Here, we show template examples with their arguments as they are actually represented for generation.
\begin{itemize}[noitemsep,topsep=2pt]
    \item [Q1:] {\bf Object verification} (54 templates)
    \begin{lstlisting}
"Does it look like there [1:Is] [1:DET] <attrs1> <obj1> anywhere?"
"[1:Is] there [1:DET] <attrs1> <obj1> anywhere?"
    \end{lstlisting}
    \item [Q2:] {\bf  Conjunctive verification} (18 templates):
    \begin{lstlisting}
"[1:Is] there both [1:DET] <attrs1> <obj1> and [2:DET] <attrs2> <obj2>?"
"Do you see both [1:DET] <attrs1> <obj1> and [2:DET] <attrs2> <obj2> in this photo?"
    \end{lstlisting}
    \item [Q3:] {\bf Disjunctive verification} (18 templates): 
\begin{lstlisting}
"Do you see either [1:DET] <attrs1> <obj1> or [2:DET] <attrs2> <obj2>?"
"Does it look like there is either [1:DET] <attrs1> <obj1> or [2:DET] <attrs2> <obj2> anywhere"
\end{lstlisting}
    \item [Q4:] {\bf Attribute verification} (25 templates):
\begin{lstlisting}
"Does it seem that the <obj1> that [1:Is] <2rel1> the <attrs2> <obj2> [1:Is] <attrs1>?"
"Would you say that the <obj1> <2rel1> the <attrs2> <obj2> [1:Is] <attrs1>?"
\end{lstlisting}
    \item [Q5:] {\bf Object multi-choice} (25 templates):
\begin{lstlisting}
"What would you call the <obj-category1> that [1:Is] <2rel1> the <attrs2> <obj2>, <obj-category-options1>?"
"Do you think the <attrs1> <obj-category1> [1:Is] <obj-category-options1>?"
\end{lstlisting}
    \item [Q6:] {\bf Attribute multi-choice} (39 templates):
\begin{lstlisting}
"Which sort of <category1> [1:Is] the <attrs1> <obj1>, <category-options1>?"
"What type of <category1> [1:Is] the <attrs1> <obj1>, <category-options1>?"
\end{lstlisting}
    \item [Q7:] {\bf Action multi-choice} (28 templates):
\begin{lstlisting}
"What kind of <category1> [1:Is] the <attrs1> <obj1> doing, <category-options1>?"
"What would you call the sort of <category1> that the <attrs1> <obj1> [1:Is] doing, <category-options1>?"
\end{lstlisting}
\end{itemize}

\begin{table*}[t]
\centering
\caption{Test dataset statistics: answer distributions for \emph{original} (non-perturbed) questions in each test dataset.}
\label{app:test_dataset_statistics}
\begin{tabular}{lccr}
\toprule
         Test Dataset & Binary & \cell{Multi-\\choice} &  Total \\\midrule
    \rephrasingtest       & 10,000 & 9,412 & 19,412 \\
    \ordertest                       &  5,000 & 9,412 & 14,412 \\
    \ontologicaltest      & 13,952 &     - & 13,952 \\
    \visualobfuscationtest & 18,000 & 8,272 & 26,272 \\
    \negationtest     & 10,000 &     - & 10,000 \\
    \attributeantonymtest &  5,000 &     - &  5,000 \\
\bottomrule
\end{tabular}
\end{table*}

\subsection{Visual Obfuscation Invariance Details}
\label{app:visual_obfuscation_invariance}
\begin{figure}[ht]
\centering \small
\begin{tabular}{c@{\hskip 0.05in}c@{\hskip 0.05in}c}
\includegraphics[height=0.7in]{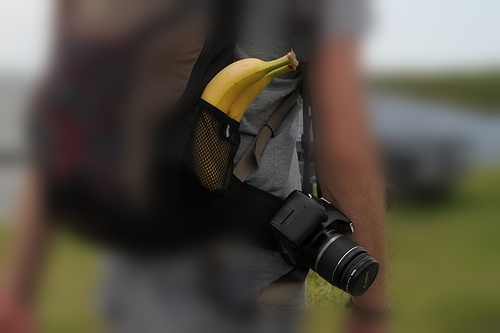} &
\includegraphics[height=0.7in]{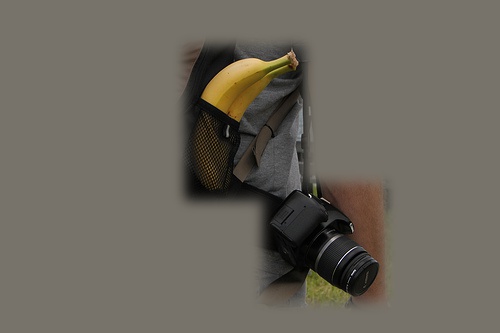} &
\includegraphics[height=0.7in]{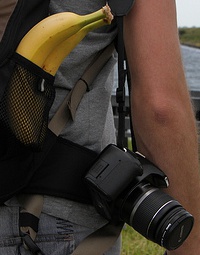} \\
Gaussian blur ($\sigma=9$) & Masking & Cropping
\end{tabular}
    \caption{Visual obfuscation  for the question ``Is there both a banana and a black camera in this photo?''}
    \label{app:context_blur}
\end{figure}

Table \ref{app:context_blur} shows examples of context blurring, masking, and cropping. Five perturbations are done in total, blurring (with $\sigma\in\{3, 6, 9\}$), masking context by replacing pixel values with the channel-wise average computed from the GQA training data, and cropping around the tightest bounding box containing the question's objects.

\begin{table*}[t]
\centering
\caption{Visual Accuracy}
\label{tab:visual_accuracy}
\begin{tabular}{lcccccc}
\toprule
            &              O &              M &              C &             G3 &             G6 &             G9 \\
\midrule
BAN            &          48.83 &          49.38 &          47.70 &          49.10 &          49.29 &          49.42 \\
CNN+LSTM       &          40.28 &          39.29 &          40.14 &          39.45 &          39.18 &          39.35 \\
LXMERT         &          69.64 &          66.33 &          61.82 &          66.59 &          66.89 &          67.06 \\
MMBT           &          50.37 &          50.00 &          49.17 &          50.44 &          50.27 &          50.18 \\
ViLBERT        &          51.60 &          51.12 &          49.99 &          51.69 &          51.36 &          51.22 \\
VisualBERT     &          53.81 &          53.25 &          52.53 &          53.67 &          53.39 &          53.44 \\
\bottomrule
\end{tabular}
\end{table*}

\subsection{Training Details}
\label{app:more_training_details_section}
We provide greater detail on training environments and hyperparameter choices. 

All models are trained using the GQA \cite{Hudson2019} balanced training set and validated on the balanced validation set (with minimal parameter tuning, aiming to stay faithful to the original implementation). All non-LXMERT models are trained and finetuned withing the MMF \cite{singh2020mmf} framework, trained using binary cross-entropy loss for a set maximum number of epochs, taking the epoch checkpoint that best performs on the validation set, and using features from Faster R-CNN \cite{Ren2015FasterRT} with a ResNext-152 \cite{resnext} backbone pre-trained on the Visual Genome \cite{krishnavisualgenome} dataset for object-based models. 

\smallsec{BAN} The Bilinear Attention Network \cite{ban} (BAN) uses pre-trained Faster R-CNN features~\cite{Ren2015FasterRT} and GloVe \cite{pennington2014glove} embeddings with an attention model and early bilinear fusion mechanism. We train ``four glimpse'' (with 4 attention heads) BAN \cite{ban} (BAN) for 13 epochs using the Adamax \cite{adam_optimizer} optimizer with an initial learning rate of $1e^{-3}$ and batch size of 256, decaying the learning rate at epochs 11 and 13. As in the original hyperparameter configuration, we perform gradient clipping at 0.25.

\smallsec{LXMERT} LXMERT~\cite{Tan2020} is a transformer-based architecture~\cite{Vaswani2017AttentionIA} with pre-trained Faster R-CNN features. It undergoes an extended pre-training procedure using 5 different pre-training tasks, including image question answering. We first pre-train a version of LXMERT from scratch with all GQA validation instances removed from the pre-training data to prevent direct leakage. We use all the default hyperparameters used in the author's GitHub repository.\footnote{\href{https://github.com/airsplay/lxmert}{github.com/airsplay/lxmert}} We then finetune LXMERT base on the GQA training dataset for 4 epochs with the same hyperparameter configuration as the original implementation, using a batch size of 32, and initial learning rate of $1e^{-5}$. LXMERT base is pre-trained using the MS COCO \cite{Lin2014MicrosoftCC} and Visual Genome datasets. LXMERT also uses a Faster R-CNN \cite{Ren2015FasterRT} with a ResNet-101 \cite{He2016DeepRL} backbone. 

\smallsec{VisualBERT} VisualBERT \cite{li2019visualbert} is similar in architecture and pre-training method to BERT~\cite{Devlin2019BERTPO}. It performs an early fusion of text and image features immediately before several transformer layers. It uses Faster R-CNN features, is initialized using weights from BERT, and pre-trained on 2 different tasks using the MS COCO dataset.  We finetune an MS COCO pre-trained version of VisualBERT using the same hyper parameters and training scheme of the original implementation for the VQA task. We use the Adam W optimizer with an initial learning rate of $2e^{-5}$ and a batch size of 64 for a maximum of 20 epochs.

\smallsec{CNN+LSTM} This model uses a 6 layer CNN module and a bidirectional LSTM module before concatenating the output of each module and passing the combined output to a FC classifier. The LSTM module uses GloVe word embeddings~\cite{pennington2014glove}. Model weights are randomly initialized. We train the model for 25 epochs using the Adam W optimizer with an initial learning rate of $1e^{-4}$ and batch size of 256. This model uses a 6 layer CNN module and a bidirectional LSTM module with a hidden size of 128, and concatenates the output of each module before passing the combined output to 2 layer MLP classifier with a ReLU activation. The LSTM module uses GloVe word embeddings \cite{pennington2014glove} to represent questions.

\smallsec{MMBT} The MultiModal BiTransformer (MMBT) \cite{kiela2019supervised} is an early fusion model, which uses Faster R-CNN features projected to a common space and concatenated with contextual BERT embeddings before being passed to transformer layers. MMBT uses pre-trained Faster R-CNN features and is initialized with BERT pre-trained weights. We finetune MMBT with the Adam W \cite{adam_optimizer} optimizer with an initial learning rate of $5e^{-5}$ with a batch size of 64, for a maximum of 15 epochs. 

\smallsec{ViLBERT} ViLBERT \cite{Lu2019ViLBERTPT} uses two parallel transformer ``streams'' for vision and language separately. These streams interact using multi-modal co-attentional transformer blocks. It uses Faster R-CNN features and is initialized using some weights from BERT. ViLBERT is pre-trained using 2 different tasks, using the MS COCO dataset. We finetune the MS COCO pre-trained version of ViLBERT in a similar manner to the finetuning scheme used for the VQA task of the original implementation. We use the Adam W \cite{adam_optimizer} optimizer with an initial learning rate of $4e^{-5}$ and batch size 64, for a maximum of 20 epochs.

\subsection{Test Results}
\label{app:more_test_results}
We provide fuller test results for each test dataset, including accuracy on the original instances and perturbed instances in tables~\ref{app:rephrasing_invariance},~\ref{app:order_invariance},~\ref{app:ontological_invariance},~\ref{app:antonym_attribute_directional_expectation}, and ~\ref{app:negation_directional_expectation}. These results supplement the primary results reported in Section~\ref{sec:results}. 
\begin{table}[ht]
\centering
\caption{Full results for \rephrasingtest.}
\label{app:rephrasing_invariance}
\resizebox{0.49\textwidth}{!}{
\begin{tabular}{lccccc}
\toprule
            &       \accuracy &  Original \accuracy &  Perturbed \accuracy &    \consistency &  \compaccuracy \\
\midrule
BAN            &          55.86 &              55.90 &           55.83 &          82.32 &                    50.20 \\
CNN+LSTM       &          37.58 &              37.61 &           37.56 &          68.93 &                    28.90 \\
LXMERT         &          67.80 &              67.70 &           67.90 &          83.83 &                    61.89 \\
MMBT           &          56.91 &              56.79 &           57.02 &          81.72 &                    50.76 \\
ViLBERT        &          56.44 &              56.54 &           56.35 &          86.44 &                    51.63 \\
VisualBERT    &          59.81 &              60.13 &           59.48 &          83.62 &                    53.87 \\\cdashlinelr{1-5}
Human          &          86.15 &              86.74 &           85.56 &          96.58 &                    85.28 \\
\bottomrule
\end{tabular}
}
\end{table}

\begin{table}[ht]
\centering
\caption{Full results for \ordertest.}
\label{app:order_invariance}
\resizebox{0.49\textwidth}{!}{
\begin{tabular}{lccccc}
\toprule
            &       \accuracy &  Original \accuracy &  Perturbed \accuracy &    \consistency &  \compaccuracy \\
\midrule
BAN            &          51.69 &              51.81 &           51.57 &          65.22 &                    38.43 \\
CNN+LSTM       &          33.25 &              33.24 &           33.26 &          62.56 &                    22.29 \\
LXMERT         &          63.26 &              63.22 &           63.30 &          71.02 &                    51.01 \\
MMBT           &          51.52 &              51.30 &           51.75 &          58.62 &                    36.65 \\
ViLBERT        &          51.02 &              51.01 &           51.04 &          66.21 &                    38.27 \\
VisualBERT    &          55.87 &              55.80 &           55.93 &          60.75 &                    40.60 \\\cdashlinelr{1-5}
Human          &          79.10 &              79.38 &           78.83 &          98.82 &                    78.76 \\
\bottomrule
\end{tabular}
}
\end{table}

\begin{table}[ht]
\centering
\caption{Full results for \ontologicaltest.}
\label{app:ontological_invariance}
\resizebox{0.49\textwidth}{!}{
\begin{tabular}{lccccc}
\toprule
            &       \accuracy &  Original \accuracy &  Perturbed \accuracy &    \consistency &  \compaccuracy \\
\midrule
BAN            &          77.67 &              79.25 &           76.10 &          77.36 &                    66.36 \\
CNN+LSTM       &          52.37 &              48.70 &           56.04 &          73.37 &                    39.06 \\
LXMERT         &          85.69 &              88.96 &           82.42 &          83.48 &                    77.43 \\
MMBT           &          73.73 &              81.84 &           65.62 &          70.46 &                    58.97 \\
ViLBERT        &          78.30 &              84.96 &           71.65 &          76.48 &                    66.56 \\
VisualBERT    &          74.82 &              80.53 &           69.10 &          74.13 &                    61.88 \\\cdashlinelr{1-5}
Human          &          96.00 &              96.00 &           96.00 &          94.00 &                    94.00 \\
\bottomrule
\end{tabular}
}
\end{table}

\begin{table}[ht!]
\centering
\caption{Full results for \attributeantonymtest.}
\label{app:antonym_attribute_directional_expectation}
\resizebox{0.49\textwidth}{!}{
\begin{tabular}{lccccc}
\toprule
            &       \accuracy &  Original \accuracy &  Perturbed \accuracy &    \consistency &  \compaccuracy \\
\midrule
BAN            &          71.02 &              81.70 &           60.34 &          56.80 &                    49.42 \\
CNN+LSTM       &          51.98 &              70.64 &           33.32 &          10.50 &                     7.24 \\
LXMERT         &          79.47 &              85.14 &           73.80 &          71.12 &                    65.04 \\
MMBT           &          69.83 &              82.74 &           56.92 &          54.58 &                    47.54 \\
ViLBERT        &          72.41 &              88.94 &           55.88 &          55.30 &                    50.06 \\
VisualBERT    &          73.76 &              82.16 &           65.36 &          59.84 &                    53.68 \\\cdashlinelr{1-5}
Human          &          87.50 &              86.00 &           89.00 &          89.00 &                    84.00 \\
\bottomrule
\end{tabular}
}
\end{table}

\begin{table}[ht!]
\centering
\resizebox{0.49\textwidth}{!}{
\caption{Full results for \negationtest.}
\label{app:negation_directional_expectation}
\begin{tabular}{lccccc}
\toprule
            &       \accuracy &  Original \accuracy &  Perturbed \accuracy &    \consistency &  \compaccuracy \\
\midrule
BAN            &          50.10 &              64.64 &           35.57 &          19.30 &                     9.73 \\
CNN+LSTM       &          49.67 &              48.51 &           50.82 &          17.85 &                     8.59 \\
LXMERT         &          52.31 &              76.08 &           28.55 &          28.11 &                    17.00 \\
MMBT           &          50.13 &              67.41 &           32.86 &          23.07 &                    11.90 \\
ViLBERT        &          49.15 &              67.58 &           30.72 &          18.73 &                     8.61 \\
VisualBERT    &          51.10 &              68.67 &           33.53 &          18.26 &                    10.23 \\\cdashlinelr{1-5}
Human          &          78.25 &              82.50 &           74.00 &          88.50 &                    74.50 \\
\bottomrule
\end{tabular}
}
\end{table}

\subsection{Model comparison results}
We provide additional model comparison results for each test dataset in Tables \ref{app:pharasl_invariance_model_coverage}, \ref{app:order_invariance_model_coverage}, \ref{app:ontological_model_coverage}, \ref{app:antonym_attr_model_coverage}, \ref{app:negation_model_coverage}.

\begin{table}[ht!]
\centering
\resizebox{0.49\textwidth}{!}{
\caption{\rephrasingtest model comparison: Entries ($C_{i, j}$) show the proportion of model $j$'s correct predictions that are also predicted correctly by model $i$. E.g. here we see that BAN correctly predicts 65.39\% of the instance pairs correctly predicted by LXMERT, but LXMERT correctly predicts 79.85\% of the correct instance pairs captured by BAN.}
\label{app:pharasl_invariance_model_coverage}
\begin{tabular}{lcccccc}
\toprule
            &           B &         C+L &           L &           M &         ViL &         Vis \\
\midrule
BAN         &           - &       75.70 &       69.42 &       75.58 &       76.88 &       73.97 \\
CNN+LSTM    &       48.58 &           - &       44.02 &       49.11 &       49.55 &       47.43 \\
LXMERT      &       83.84 &       80.09 &           - &       83.23 &       84.40 &       83.01 \\
MMBT        &       76.53 &       76.22 &       69.77 &           - &       79.47 &       77.58 \\
ViLBERT     &       77.00 &       76.27 &       70.12 &       78.74 &           - &       76.87 \\
VisualBERT &       78.95 &       76.93 &       73.39 &       81.89 &       81.80 &           - \\
\bottomrule
\end{tabular}
}
\end{table}

\begin{table}[ht!]
\centering
\caption{Model coverage results for \ordertest.}
\label{app:order_invariance_model_coverage}
\resizebox{0.49\textwidth}{!}{
\begin{tabular}{lcccccc}
\toprule
            &           B &         C+L &           L &           M &         ViL &         Vis \\
\midrule
BAN         &           - &       76.20 &       67.90 &       75.41 &       76.28 &       72.81 \\
CNN+LSTM    &       46.63 &           - &       41.31 &       47.71 &       47.94 &       45.95 \\
LXMERT      &       82.49 &       78.80 &           - &       82.10 &       82.86 &       81.14 \\
MMBT        &       74.95 &       75.90 &       67.06 &           - &       78.11 &       76.10 \\
ViLBERT     &       74.88 &       75.72 &       67.02 &       77.31 &           - &       74.11 \\
VisualBERT &       78.42 &       78.36 &       71.91 &       82.56 &       81.12 &           - \\
\bottomrule
\end{tabular}
}
\end{table}

\begin{table}[ht!]
\centering
\caption{Model coverage results for \ontologicaltest.}
\label{app:ontological_model_coverage}
\resizebox{0.49\textwidth}{!}{
\begin{tabular}{lcccccc}
\toprule
            &           B &         C+L &           L &           M &         ViL &         Vis \\
\midrule
BAN         &           - &       81.71 &       81.91 &       84.95 &       83.92 &       83.85 \\
CNN+LSTM    &       55.09 &           - &       53.05 &       54.57 &       53.64 &       53.70 \\
LXMERT      &       90.36 &       86.81 &           - &       91.20 &       91.81 &       90.89 \\
MMBT        &       80.64 &       76.83 &       78.48 &           - &       82.13 &       83.12 \\
ViLBERT     &       84.60 &       80.21 &       83.90 &       87.22 &           - &       88.14 \\
VisualBERT &       80.77 &       76.71 &       79.36 &       84.34 &       84.22 &           - \\
\bottomrule
\end{tabular}
}
\end{table}

\begin{table}[ht!]
\centering
\caption{Model coverage results for \attributeantonymtest.}
\label{app:antonym_attr_model_coverage}
\resizebox{0.49\textwidth}{!}{
\begin{tabular}{lcccccc}
\toprule
            &           B &         C+L &           L &           M &         ViL &         Vis \\
\midrule
BAN         &           - &       79.06 &       78.60 &       82.55 &       82.89 &       82.21 \\
CNN+LSTM    &       57.94 &           - &       55.13 &       59.96 &       59.20 &       57.48 \\
LXMERT      &       87.98 &       84.30 &           - &       87.19 &       88.06 &       87.63 \\
MMBT        &       81.07 &       80.03 &       76.56 &           - &       83.04 &       82.15 \\
ViLBERT     &       84.48 &       82.22 &       80.20 &       86.21 &           - &       84.84 \\
VisualBERT &       85.38 &       81.47 &       81.32 &       86.88 &       86.45 &           - \\
\bottomrule
\end{tabular}
}
\end{table}

\begin{table}[ht!]
\centering
\caption{Model coverage results for \negationtest.}
\label{app:negation_model_coverage}
\resizebox{0.49\textwidth}{!}{
\begin{tabular}{lcccccc}
\toprule
            &           B &         C+L &           L &           M &         ViL &         Vis \\
\midrule
BAN         &           - &       55.26 &       66.06 &       68.95 &       71.67 &       68.69 \\
CNN+LSTM    &       54.73 &           - &       52.40 &       54.11 &       54.66 &       53.57 \\
LXMERT      &       68.91 &       55.16 &           - &       70.95 &       72.17 &       70.60 \\
MMBT        &       68.99 &       54.64 &       68.08 &           - &       73.13 &       71.83 \\
ViLBERT     &       70.24 &       54.08 &       67.76 &       71.69 &           - &       73.48 \\
VisualBERT &       70.04 &       55.16 &       68.96 &       73.19 &       76.43 &           - \\
\bottomrule
\end{tabular}
}
\end{table}

\end{document}